\documentclass{article}

\usepackage[round]{natbib}
\usepackage[preprint]{neurips_2025}

\usepackage[utf8]{inputenc} % allow utf-8 input
\usepackage[T1]{fontenc}    % use 8-bit T1 fonts
\usepackage{hyperref}       % hyperlinks
\usepackage{url}            % simple URL typesetting
\usepackage{booktabs}       % professional-quality tables
\usepackage{amsmath}        % for \operatorname and math environments
\usepackage{amsfonts}       % blackboard math symbols
\usepackage{nicefrac}       % compact symbols for 1/2, etc.
\usepackage{microtype}      % microtypography
\usepackage{xcolor}         % colors
\usepackage{blindtext}
\usepackage{hyperref}
\usepackage{graphicx}

\title{Neural Inhibition Improves Dynamic Routing and Mixture of Experts}

\author{%
  Will Y. Zou \\
  Angle.ac\\
  San Francisco, CA, 94131 \\
  \texttt{will@angle.ac} \\
  \And
  Jennifer Y. Zhang \\
  Engineering Science\\
  University of Toronto \\
  Toronto, ON, CA M5S 1A1\\
  \texttt{jenniferyt.zhang@mail.utoronto.ca} \\
}

\begin{document}

\maketitle

\begin{abstract}
To be effective, efficient, and diverse, deep learning models need to dynamically choose its architecture based on signals from a population of neurons. We hypothesize dynamic routing models can be improved with neural inhibition in those neural populations. This means signals commonly shared among the various modes of data statistics can be inhibited so that the routing model can choose a specialized expert path for each data sample. Only through inhibition is the routing mechanism able to effectively select neural pathways. We believe this is an under-studied and under-verified implementation methodology for Mixture-of-Experts, dynamic routing, and transformer language models. We provide experimental evidence that the neural inhibition algorithm significantly boosts the performance of general tasks and motivates more effort to be invested in this research direction. 
\end{abstract}

\section{Introduction}
\label{sec:intro}
The ability to adaptively and dynamically change architecture makes machine learning models powerful and efficient. When a model performs an important and general task, it often encounters data of multiple types. In such cases, it is beneficial for multiple experts to handle the task by focusing on selective features identified in the data.

We hypothesize that neural inhibition mechanisms, inspired by biological neurons, can support these routing decisions, enabling the model to become more specialized and robust. To achieve this, we need to establish the computational basis for selectively inhibiting signals during decision-making. Inhibition serves as a computational primitive for this selective suppression. We argue that inhibition within the input representation space can enhance dynamic routing mechanisms, such as those used in Mixture-of-Experts models, by improving specialization and efficiency.

In the biological cortex, inhibitory connections help modulate neuron excitatory activity. They control the flow of information to maintain network stability. Reviews of sensory cortex show that lateral and feed-forward inhibition sharpen selectivity and signal-to-noise ratio of downstream neurons \cite{Isaacson2011231}. Studies show inhibition exhibits remarkable diversity in terms of morphology, molecular markers, electro-physiological properties, and connectivity patterns \cite{klausberger2008neuronal,gidon2012principles,markram2004interneurons,Chini2022InhibitionDecorrelation,sale2010gabaergic}. 

For deep learning, this computational basis of inhibition has been imported in the form of learnable gating mechanisms \cite{hochreiter1997long,dauphin2017languagemodelinggatedconvolutional} which can be referred to as ``\textit{inhibition gating}''. They are parametric modulators whose outputs can amplify and suppress features. Applied in Long Short-Term Memory (LSTM), Gated Linear Units (GLU)~\cite{shazeer2020gluvariantsimprovetransformer}, the architecture helps units learn important representations of the input while enhancing network sparsity. \cite{xie2000} applies winner-take-all (WTA) competition between neurons by lateral inhibition. Other works investigate sparse coding and locally competitive networks \cite{rozell2008sparse,shapero2014optimal,bahadi2024efficientsparsecodingadaptive} which defines inhibition strength for more robust representations of input data to solve various downstream tasks. It is important to note that a key property of ``\textit{inhibition gating}'' is its flexibility in source and target connectivity. The inhibitory signals can arise from any neuron and any layer and be directed to any subset of the network. The above computational models all demonstrate such diverse inhibition, drawing parallels to observations in the mammalian cortex \cite{klausberger2008neuronal, Isaacson2011231, tremblay2016gabaergic}. 

Moreover, Long Short-Term Memory networks introduce sigmoidal input, forget gate, and output gate, to prevent vanishing gradients while filtering out irrelevant history \cite{hochreiter1997long, lstm}.  Highway Networks generalize the idea of arbitrarily deep feed-forward networks through transform gates and carry gates that learn to regulate the flow of information through the network \cite{SrivastavaGS15}. \cite{hua2019channelgatingneuralnetworks} applies channel gating to convolutional neural networks to identify features that are useful to perform downstream vision tasks. \cite{dauphin2017languagemodelinggatedconvolutional} introduces a gated convolutional network to capture long-term dependencies of unbounded text inputs. Gated Linear Units (GLU) are introduced in the paper as a gating mechanism to effectively and efficiently filter out unhelpful features by providing a linear path for gradients while retaining non-linear capacity. Later, \cite{shazeer2020gluvariantsimprovetransformer} empirically shows that incorporating a GLU-based feedforward network in a Transformer-based model improves the model's ability to perform various standard language tasks. 

% For dynamically routing data in neural architectures, multiple types of data exist in the input with different statistics. A mixture of experts could improve the capability of the model when the model is asked to perform a general task on all data distributions. The dynamic routing system assumes that the optimal neural architecture is different for each data distribution. A stereotypical example is A Mixture of Experts model. The model may dynamically route data to an expert based on features characterizing the data. 

Despite the success of ``\textit{inhibition gating}'' mechanisms, state-of-the-art dynamic-routing and Mixture-of-Experts (MoE) models have not inherited this principle. There is no explicit mechanism to suppress irrelevant routing signals at the input space, and the models suffer from various problems. For example, \cite{shazeer2017outrageouslylargeneuralnetworks} pioneers a top-$k$ softmax router with added noise to spread the load, achieving huge capacity gains but at the cost of complex routing logic. \cite{fedus2022switchtransformersscalingtrillion} then introduces Switch Transformer to show that picking just one expert per token achieves great performance and reduces memory requirements, but it can suffer from imbalanced expert utilization. \cite{zhou2022mixtureofexpertsexpertchoicerouting} flips the paradigm by letting each expert choose which inputs to process, improving load balancing while similar MoE architectures continue to observe instability. Most recently, \cite{jiang2024mixtralexperts} proposes a Sparse Mixture of Experts (SMoE) on a large language model where each feedforward sub-block is replaced by an MoE layer containing eight experts and a softmax-based router. However, the model is computationally heavy, and it retains a load-balancing loss where if this coefficient drifts, training can destabilize. 

We hypothesize that ``\textit{inhibition gating}'' mechanisms are critical to aid the pathway decisions in dynamic routing networks. We aim to bridge the gap between inhibition mechanisms and the current lack of its analog in dynamic routing networks. We advocate for exploring a \textit{globally-connected inhibition} mechanism where any neuron can inhibit any other. 

To verify this hypothesis, we propose a methodology to analyze the computational basis of inhibition through simulation. We limit the scope of this paper to the application of inhibition and studying its role in dynamic routing architectures of artificial neural networks. We study the scope of problems that satisfy two conditions: first, the machine learning problem has multiple types of data statistics in the input, towards each of which a specific network could act as an expert; second, the model should be performing a generic task, and it is not specific to one type of input. Within these conditions, we believe that inhibition of the population of neurons used for routing is crucial and can significantly improve the capability of the global model. Moreover, we present experiments and examples to further support our claim. We empirically validate our position that biologically inspired global inhibition mechanisms help make dynamic routing decisions.

\textbf{Paper Structure.} Section \ref{sec:method_part} presents our research scope and method for supporting our hypothesis. Section \ref{sec:exp_part} describes empirical results to illustrate the critical role of inhibition in dynamic routing. Section \ref{sec:future} proposes several interesting future directions to be explored. Finally, Section \ref{sec:conclusion} contains final remarks. 

\section{Related Work}

A similar but less biologically plausible inhibition method is the concept of ``dropout'' introduced by \cite{hinton2014}. The essence is to incorporate a probability that defines the portion of neurons that will be deactivated during the network training. However, ``dropout'' uses a fixed mask rate, which cannot adapt to layer importance. Therefore, \cite{frey2013} and \cite{kingma2015variationaldropoutlocalreparameterization} both introduce adaptive dropout, proposing a learnable mask rate, so that more important neurons are less likely to be deactivated. Moreover, independent neuron dropout ignores spatial correlation in convolutional feature maps when solving vision tasks. Therefore, \cite{ghiasi2018dropblockregularizationmethodconvolutional} created ``DropBlock'' where neurons in a region of a feature map are dropped together for convolutional neural networks. In recent years, as the popularity of Transformer-based models increases, the ``dropout'' mechanism is applied to vision transformers \cite{liu2022patchdropouteconomizingvisiontransformers,zhao2022revisitingstructureddropout,li2023dropkey}.

Mixture-of-Experts (MoE) was first introduced by \cite{hinton1990} to show that a gating function, commonly known as a ``router'', can assign tasks to local expert models and specialize them via gradient updates. It has gained extensive popularity in the past few years scaling to deep networks, Transformer-based models, and massive large models. \cite{shazeer2017outrageouslylargeneuralnetworks} pioneers a top-$k$ softmax router with added noise to spread the load, achieving huge capacity gains but at the cost of complex routing logic. \cite{fedus2022switchtransformersscalingtrillion} then introduces Switch Transformer to show that picking just one expert per token achieves great performance and reduces memory requirements, but it can suffer from imbalanced expert utilization. \cite{zhou2022mixtureofexpertsexpertchoicerouting} flips the paradigm by letting each expert choose which inputs to process, which improves load balancing and stability. Most recently, \cite{jiang2024mixtralexperts} builds on previous work and proposes a Sparse Mixture of Experts (SMoE) on a large language model where each feedforward sub-block is replaced by a MoE layer containing eight experts and a softmax-based router. It demonstrates great performance and further confirms the significance of MoE architecture on large Transformer-based models.

Our work brings together the above ideas from gating, biological networks, dropout, and MoE to propose a global, data-driven inhibition mechanism that operates at the level of all layers and even the whole network. Unlike dropout, which randomly turns off units, our adaptive inhibition unit learns, via a small secondary network, a continuous mask over neurons in a set that dynamically scales each activation inhibit signal. This goes beyond classic sparse-coding/LCA approaches where lateral inhibition is local and threshold-based— allowing all-to-all interactions, even from later layers. We produce smooth and soft network gating rather than hard WTA spikes. While traditional MoE models route the input to various expert models in parallel, our inhibition mechanism learns to direct tasks to experts at multiple layers to enhance the overall decision-making process. Moreover, we discuss and illustrate the effectiveness of the inhibition mechanism on multiple downstream tasks such as vision classification and language tasks. Our work provides a novel hypothesis and verifies the hypothesis through systematic experiments, and provides a broader and more updated insight compared with prior work. 

Importantly, in the backdrop of investigations in Mixture-of-Experts, Dropout, and Neuroscience, we argue it is vital to recognize the effective global inhibition techniques for scalable deep learning models. \textbf{We take the position that the state-of-the-art deep learning models can be advised by global inhibition techniques}, given we show evidence of its potential. 

% \section{Why Inhibition in Dynamic Routing}
% \label{sec:inhibition_in_DR}

\section{Research Design and Methodology} 
\label{sec:method_part}
Despite previous attempts to adopt adaptive methods for routing architectures from the neuron level~\cite{frey2013,wan2013regularization}, it remains unclear whether the adaptive techniques can improve deep learning models at scale. One such opportunity is the growing interest in Mixture-of-Experts (MoE) which is proven to scale model capacity for state-of-the-art large language models while maintaining computational efficiency. Despite their promise, these models continue to require supporting research~\cite{zhou2022mixture,dai2024deepseekmoe,shazeer2017outrageouslylargeneuralnetworks, fedus2022switchtransformersscalingtrillion, zhou2022mixtureofexpertsexpertchoicerouting, jiang2024mixtralexperts}, sometimes heuristics, to resolve challenges including expert over-utilization and balancing, and training instability. This is likely due to the optimization search space being modified by dynamic routing, while the model aims to solve generic tasks without meta-labels to help route data to experts.

With various background studies on biological neurons, we argue that neural inhibition is one of the computational principles that can play a critical role in dynamic routing neural networks. In biological neural circuits, inhibitory neurons exhibit a broad diversity in terms of connections and morphology. The inhibitory connections control excitatory flow and maintain stability, improve selectivity, and promote sparsity \cite{wilson2012division, froemke2015inhibition, urban2020diverse, udakis2020interneuron, hage2022circuit}. For their computational basis in artificial networks, these properties are translated to ``\textit{inhibition gating }'', allowing any-to-any neuron inhibition in the network. Inspired by this, we provide evidence of the effectiveness of a diverse set of inhibition mechanisms with MoE models, and take the position this should be widely researched as a common practice for dynamic routing models. 

As a motivation for our research design, we investigate the problem where a generic task is performed on input data, and the input exhibits multiple types or modes of data statistics. Here a Mixture-of-Experts model can potentially deal with the multiple types of data. Without diverse and global inhibition, we hypothesize that it's challenging for routing models to correctly direct the data. 

We propose a research design with the following conditions, to investigate the problem in question: 
\begin{enumerate}
\item \textbf{Different Types of Input.} The input data must contain more than one distinct type. These types are not annotated and they act as factors that a MoE router should learn to separate on its own.
\item \textbf{Generic Task.} The tasks we assign to the model are generic. They are not classifying the modes of data statistics present in the input, but rather, it is a task that's orthogonal to it. 
\item \textbf{Scaling Potential.} The enlarging or scaling of the model with   Mixture-of-Experts will improve the results of the generic task. 
\end{enumerate}

In this setting, we formalize our position through the design and validation using experiments. We posit with our position, that ``\textit{inhibition gating}'' on the population or neurons for routing will boost performance on general tasks. The experiments will provide evidence for the inhibitory connections that would selectively encourage routing models to distinguish different unlabeled data types. Before that, it's important to describe our methodology to address the characteristics of the ``\textit{diversity}'' of inhibitory neurons in biological networks.

\subsection{Diversity of Inhibition Network Connections}
\label{subsec:inhibition_methods}
The diversity in biological inhibitory neurons means there is a diversity of morphologies and connections. In a computational context, we focus on the diversity of connections as the mathematical formulations allow us to simulate other types of diversity with wide connectivity. 

We argue that in the early works of Mixture-of-Experts for computational models, the connections are not diverse as the router model only connects to the input feature layer. To simulate the diversity of connections, we adopt a variety of connection models that allow connections from any part of the neural network to the input of the routing model of the MoE. These connections are connected to the input layer through \textit{sigmoid gating}, similar to those used in LSTMs~\cite{hochreiter1997long} and GLUs~\cite{dauphin2017languagemodelinggatedconvolutional}. 

% We illustrate the merits of the algorithms from a theoretical perspective.

% This is done through better denoising the signal with which the routing model choose the expert model or path for the seen data. 

% \textbf{Random Inhibition.} The first reference algorithm is simply randomly silencing neurons in the population. This is implemented with dropout by \cite{hinton2014}. This simulates inhibition of neural signals without connections with other neurons in the network.
\textbf{Random Inhibition}. There are already techniques for randomly applying inhibition to units. This can be done with Dropout. 

\textbf{Single Cross-layer Inhibition.} We apply the sigmoid function to the out of an extra linear fully-connected layer, to the layer of neurons before the MoE router. This is the same as the formulation of Gated Linear Units~\cite{dauphin2017languagemodelinggatedconvolutional}. We element-wise multiply the pre-activation $z$ with the output of the sigmoid activation function $\sigma$ on input $x$:  
\begin{equation}
    z^\star = z \odot  \sigma (\mathbf{w}^T x + b).
\end{equation}

\textbf{Global Inhibition with Pre-Text Connections.}
For the inhibitory model to be adaptive, we propose pre-text connections from various parts of the model before the current layer:
\begin{equation}
z^\star = z \odot  \sigma (G(x) + \sum_iPr_i(x_i)) .
\end{equation}
Here $G$ is the gated linear unit extra layer, and $Pr_i$ is the pre-text network, whose input $x_i$ is the output of neurons in earlier parts of the network. Index $i$ denotes the multitude of connections from earlier in the network, where the sequence is determined by forward propagation.

\textbf{Global Inhibition with Post-Text Connections.} 
To illustrate the diversity of connections with inhibitory neurons, we propose to simulate the backward connections from the later part of the network, even the loss layers. These connections will provide highly informative signals to inhibit signals at the MoE routing layer. The challenge here is to avoid the potential introduction of loop circuitry or unwanted recurrent connections. To address this, we propose a novel algorithm to work with stochastic or mini-batch optimization algorithms for networks. 

The implementation leverages PyTorch’s forward hooks to intercept and store activations without propagating gradients. In the subsequent iteration, a forward pass through the inhibition prediction model generates the inhibition activations for the current batch, and the gradients are computed based on the loss from the batch. To address inconsistencies from varying batch sizes and unmatched samples between iterations, we apply a max-pooling operation across the training batches. This ensures that the inhibition activations can be broadcasted to all samples in the current batch. The following equation shows the Post-text connection model and $k$ denotes the optimization iterations. 

\begin{equation}
    z^{\star (k)} = z^{(k)} \odot  \sigma (G(x^{(k)}) + \sum_j \operatorname{maxpool}(Po_j(x_j^{(k-1)}))) .
\end{equation}

\textbf{Global Inhibition Model for `All-to-Local' Inhibition.} 
We combine all three approaches above to arrive at the Global Inhibition Model:

\begin{equation}
z^{\star (k)} = z^{(k)} \odot  \sigma (G(x^{(k)}) + \sum_iPr_i(x_i)) + \sum_j \operatorname{maxpool}(Po_j(x_j^{(k-1)}))) .
\end{equation}

The Global Inhibition Model implements single-layer, pre-text, and post-text connections for inhibiting the local layer of the MoE router. This is an 'All-to-Local' connection for inhibitions from all other neurons in the network to the local layer. 

\textbf{All-to-All Connections.} We intend as part of future  work, we are to investigate the `All-to-All' inhibition model. This is a model where all neurons have inhibitory connections to all neurons in the network. 

\section{Validation Experiments}
\label{sec:exp_part}
We validate our position through two experiments on vision classification task and language model word prediction task. The results for both experiments validate our position that global inhibition is effective in improving model performance on generic tasks.

\subsection{Hand-written Digits and Number of Squares Experiment} 
\label{subsec:numbers_data}

\textbf{Objective.} Our objective is for our proposed model to learn to distinguish hand-written digits and synthetic square patterns while being able to classify given hand-written digits correctly. We train a vision recognition Mixture-of-Experts (MoE) model whose router must first decide which modality it is seeing, whether it is hand-written digits or synthetic square patterns, and then assign the sample to the most suitable experts. The ultimate goal is to correctly classify the digits as $0-9$. 

\begin{figure*}[htbp]
    \centering
    \includegraphics[width=0.5\linewidth]{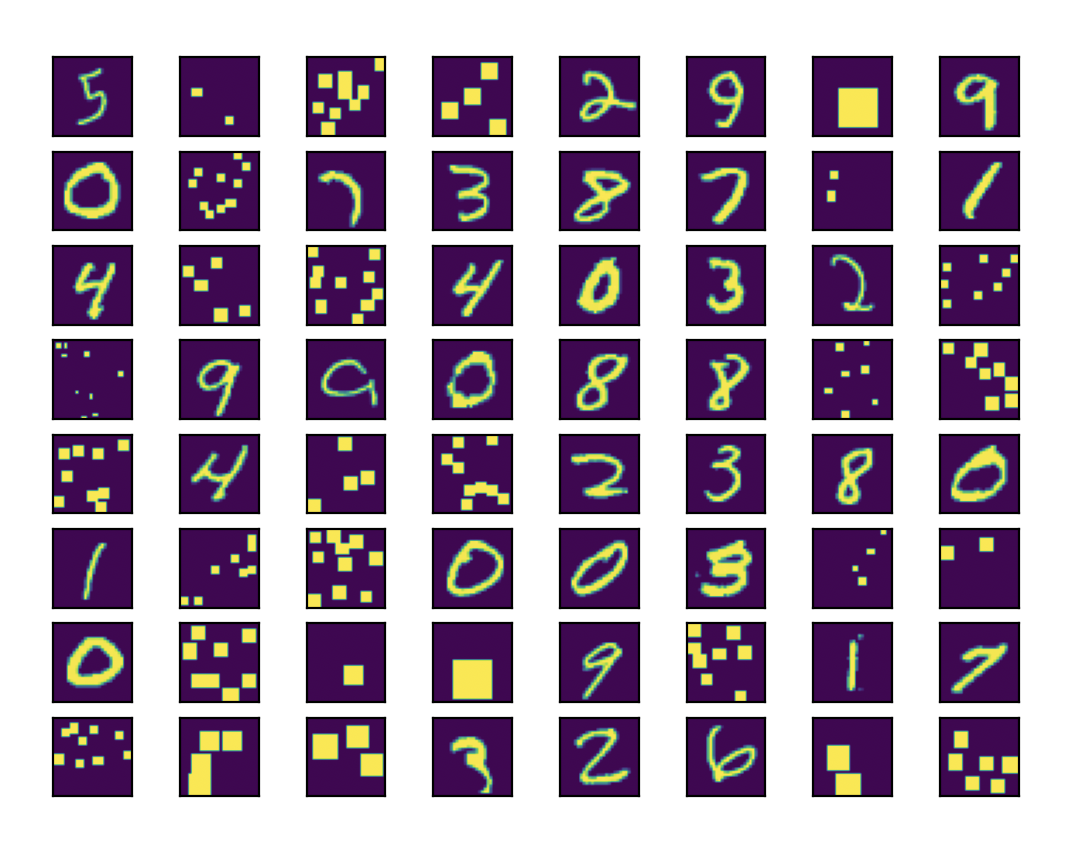}
    \caption{Examples from the mixed-numbers dataset with hand-written digits and number of squares.}
    \label{fig:numbers_dataset}
\end{figure*}

\textbf{Dataset.} We first construct a new dataset with multiple modes of data statistics. This dataset contains $120,000$ samples of $28\times28$ pixels visual fields. $60,000$ are hand-written digits from $0$ to $9$ from MNIST training dataset. $60,000$ are number of squares following \cite{stoianov2012emergence}. Each sample could either be a hand-written digit, or it could be several squares of various sizes. The supervised label for each sample is the number (0-9) present in the input. We do not use explicit labels of whether the sample is a hand-written digit or a set of squares. The dataset satisfies conditions proposed in Section \ref{sec:method_part}: it contains two distinct types of data statistics and the task is generic, not correlated with the data. Examples from this dataset are shown in Figure~\ref{fig:numbers_dataset}. 

% Using the above mixed-numbers dataset, we train a vision recognition Mixture-of-Experts (MoE) model whose router must first decide which modality it is seeing, whether it is hand-written digits or synthetic square patterns, and then assign the sample to the most suitable experts. The ultimate goal is to correctly classify the digits as $0-9$. 

% The mixed-numbers dataset uses the MNIST training set of 60,000 samples. We use an automated script to generate another 60,000 samples with number of squares. Across these 120,000 samples, we use 80\% as training, 10\% as validation and 10\% as test. The generation of the squares data follows~\cite{stoianov2012emergence}.

\textbf{Model Architecture.} The baseline model architecture uses a two-layer convolution network, interleaved with max-pooling. On top of this model, we construct a sparsely-gated Mixture-of-Expert model as described in~\cite{shazeer2017outrageouslylargeneuralnetworks}. For each expert, we employ a two-layer fully connected network with $128$ hidden units. The router applies a two-layer fully connected network with 128 hidden units. For each data point, we choose $K=3$ experts out of $5$ experts. 

\textbf{Training.} To optimize the model, we employ the Adam optimizer with a learning rate of $5 \times 10^{-5}$. For training, we use a mini-batch size of $128$ and train for $35$ epochs across the dataset. For evaluation, we use a test batch of $5,120$, and performance is averaged over 5 random initializations. We randomly shuffle the dataset and split the data with $80\%$ for training, $10\%$ for validation, and $10\%$ for testing. 

\textbf{Result.} The result of the experiment is shown in Table~\ref{tab:results_mnist_squares}. It summarizes the test accuracy on the Mixed-numbers benchmark. We observe that the Mixture-of-Expert model indeed requires a form of de-noising or inhibition on the input to perform well. The random inhibition using dropout is improving performance. However, the choice of the dropout ratio requires tuning for hyper-parameters. While the adaptive inhibition methods can automatically determine the level of signal adjustments in the neural population. Furthermore, the more diverse the in-coming connections, or, the more global the inhibition model, the better the model performs on the generic task of recognizing numbers from the input. 

\begin{table}[htbp]
\parbox{.50\linewidth}{
  \centering
  \caption{Number recognition: Mixed-numbers dataset.} 
  \label{tab:results_mnist_squares}
  \begin{tabular}{lll}
    \toprule
    Algorithm & Acc. & Std.\\ 
    \midrule 
    Baseline model & 81.4\% & 1.7e-2 \\
    \midrule 
    MoE 3/5 experts & 92.3\% & 5.6e-2\\
    Random (dropout 0.25) &94.9\% & 2.0e-2 \\
    Random (dropout 0.5) & 95.1\% & 1.9e-2  \\
    Random (dropout 0.75) & 95.9\% & 8.1e-3 \\
    One-layer I. (GLU) & 95.5\% & 6.8e-3\\
    Pre-text Inhibition  & \bf{96.6\%} &2.3e-3\\
    Global Inhibition  &  \bf{96.7\%} & 3.4e-3\\
    \bottomrule
    \end{tabular}
}
\parbox{.50\linewidth}{
  \centering
  \caption{Normalized log-likelihood loss: test split of WMT English monolingual dataset} 
  \label{tab:results_llm}
  \begin{tabular}{lll}
    \toprule
    Algorithm & 300k data & 1m data \\ 
    \midrule 
    Baseline model & 3.69e-6 & 1.17e-8\\
    \midrule 
    MoE 3/10 experts & 3.68e-6 & 2.52e-10\\
    One-layer I. (GLU) & 3.37e-6 &1.82e-10\\
    Global Inhibition &  \bf{3.31e-6} &\bf{1.81e-10}\\
    \bottomrule
    \end{tabular}
}

\end{table} 

%\subsection{Attention routing in visual QA} 

%In this sub-section, we shift gears to consider a visual fixation model. Instead of routing data to model experts, this scenario investigates how visual fixation can be decided on an image. The shared routing model sees the entire image, while combined with language prompts, it must decide which area in the image the main model should fixate on. With the fixation decisions, the visual model is applied to answer questions correctly. We experiment with adaptive inhibition on the shared routing visual model. 

\subsection{Language Model Experiment}

\textbf{Objective.} Our goal is to maximize the language model’s next-word prediction accuracy across text sequences. That is, given the preceding context of tokens, the model should accurately estimate the conditional probability distribution over the vocabulary and select the most plausible continuation. We evaluate this via normalized log-likelihood. The lower the normalized log-likelihood, the better the model performs.

\textbf{Dataset and Data Processing.} We experiment with two subsets of the WMT monolingual English datasets by \cite{maillard-etal-2024-findings}. We pre-process the data in the following way: all texts are lower-cased and porter-stemmed; number digits are replaced with \texttt{<num>}; contractions are expanded (e.g. replacing "can't" with "can not"); punctuations are removed. The smaller dataset contains $300,000$ sentences and $7,481,087$ words, and the larger dataset contains $1,000,000$ sentences and $24,968,036$ words. 

\textbf{Model Architecture.} We construct a Transformer-based language model embedded with Mixture-of-Experts. The baseline model applies a two-layer transformer with embedding dimensions $50$, hidden dimensions $50$, and number of heads $2$. We leverage the `nn.Transformer' class in pytorch. We mainly use a decoder architecture for the transformer and construct a customized decoder layer for the sparsely-gated MoE architecture with fully-connected networks as experts. This replaces the fully connected network after the self-attention block. In more detail, the embeddings are routed to multiple experts as described in \cite{shazeer2017outrageouslylargeneuralnetworks}~ and \cite{du2022glam}, without expert-balancing regularization or techniques. 

\textbf{Training.} For the smaller dataset with use a language model with a smaller vocab size of $8,500$ most frequently seen words, and for the larger dataset, we use a vocabulary size of $15,000$ most frequently seen words. For optimization, we use the Adam optimizer with a learning rate of $0.001$ and a batch size of $256$ for training and $5,120$ for testing. For both small and large datasets, we split the data $80\%$ for training,$ 10\%$ for validation, and $10\%$ for testing. 

\textbf{Result.} Table~\ref{tab:results_llm} shows the performance of the language model on word prediction. We measure the test normalized log-likelihood (nll) after training for $3$ epochs. It can be shown that adaptive inhibition significantly boosts performance in the scalable language model experiments. We observe that without adaptive inhibition, the $3/10$ (choosing $3$ experts out of $10$) MoE model performance is almost on par with the baseline model. With global inhibition, we observe the performance improves even further for both datasets. 

\section{Analysis and Interpretation} 

With our proposed methodology explained in Section~\ref{subsec:inhibition_methods}, we analyze the feature embeddings at the input to the Mixture-of-Experts (MoE) router. The goal of this analysis is to validate the hypothesis that the inhibition mechanism we introduce is capable of selectively suppressing both "common" and "discriminative" features present in the population of neurons. This suppression facilitates more effective routing of inputs to specialized expert sub-networks, allowing better model performance. Our analysis is based on the mixed-numbers dataset proposed in Section \ref{subsec:numbers_data}.

\textbf{Pearson Correlation.} We seek to understand the correlations between the feature embeddings $\mathbf{f_i}$ for the MoE router, and the data statistic types $\mathbf{t}$ in the input. In other words, the measure tells which neurons' activations are indicative of the type of input. For the $i$-th feature for the router, we calculate the $c_i$ measure for how much the feature is indicative of the data type, e.g. for the numbers dataset, whether it is a digit or a set of squares.
\begin{equation}
c_i = \frac{\mathbf{E}[(f_i-\overline{f_i})(t - \overline{t})]}{\sigma_{f_i}\sigma_{t}}.
\end{equation}

In the mixed-numbers dataset, we annotated the type of number as a meta-label, and for each of the $128$ neurons in the router input, we computed the Pearson correlation between that neuron's activations and the type meta-label. We then take the absolute value since negative correlation can be leveraged to discriminate the type of numbers. If this negative correlation is lower, it indicates that the feature is a `common' characteristic across the two types of data. If it is higher, it represents that the feature can help to `discriminate' the two types of data. We use the mean of the Pearson Correlation as the decision boundary. 

\textbf{Inhibition Thresholds.} To deepen our analysis, we evaluate a finely spaced range of inhibition thresholds to quantify the number of "common" features and "discriminative" features that are inhibited.  This allows us to build a plot of the number of inhibited common features against the number of inhibited discriminative features.
\begin{figure*}[htbp]
    \centering
    \begin{minipage}{0.45\textwidth}
        \centering
        \includegraphics[width=0.9\textwidth]{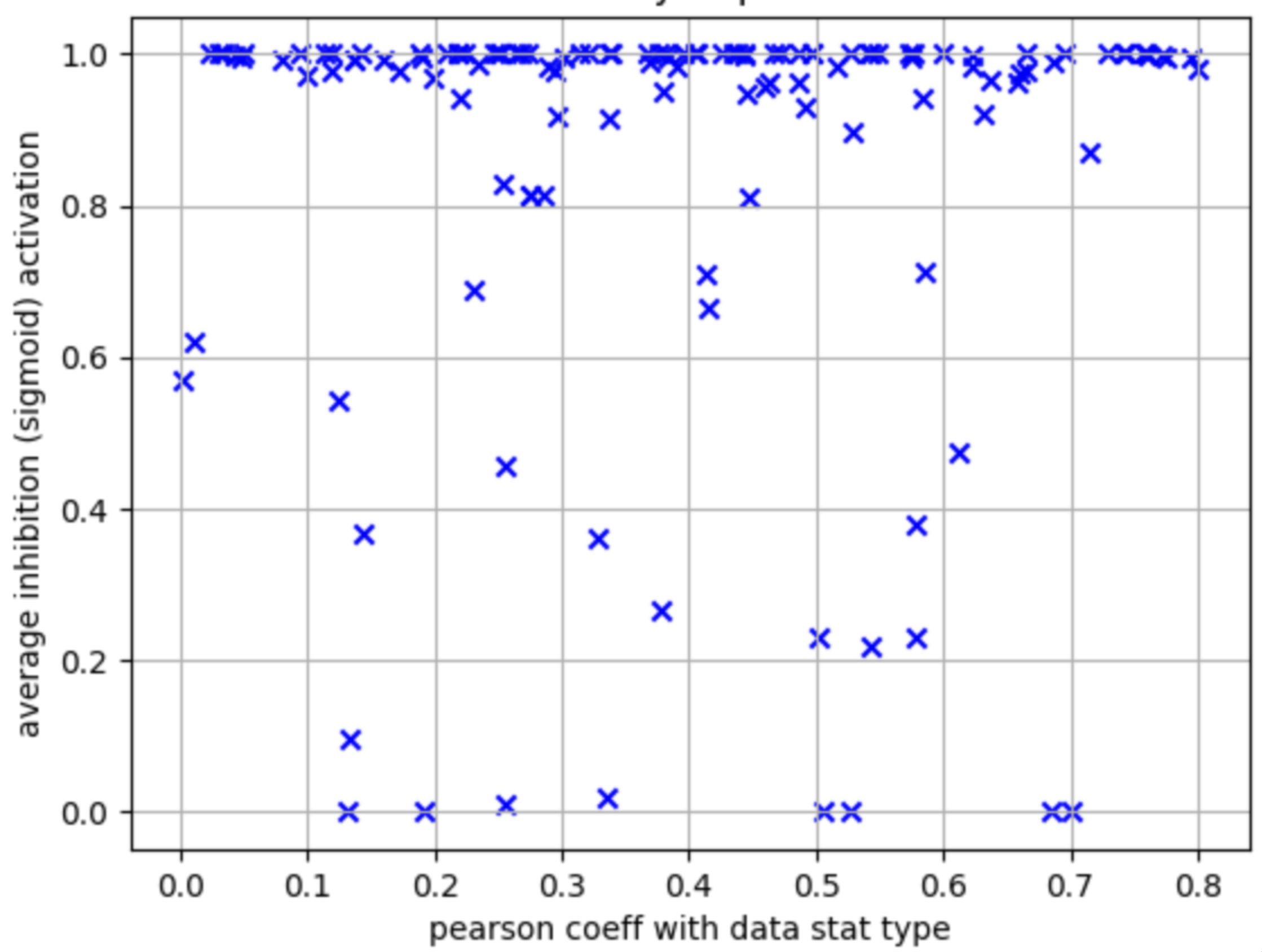} % first figure itself
        \caption{Pearson correlation vs average inhibition activations.}
        \label{fig:inhibition_vs_p_corr_plot}
    \end{minipage}\hfill
    \begin{minipage}{0.45\textwidth}
        \centering
        \includegraphics[width=0.9\textwidth]{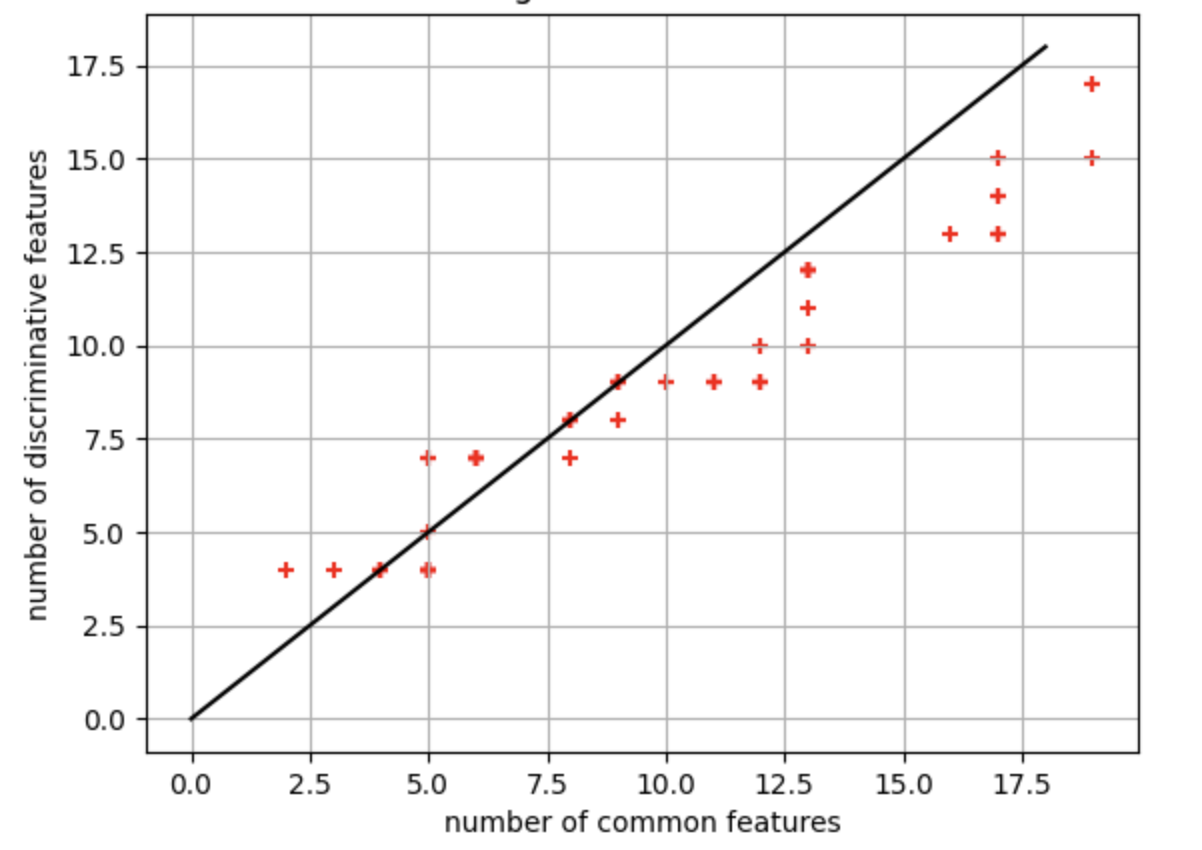} % second figure itself
        \caption{Number of common vs discriminative features with inhibition thresholds.}
        \label{fig:common_vs_dis}
    \end{minipage}
\end{figure*}

Figure~\ref{fig:inhibition_vs_p_corr_plot} plots the average inhibition activation of each neuron versus its absolute Pearson correlation between neuron activation and the type of number data (digit or square). It is observed there are a number of neurons receiving inhibitory suppression, even near zero average activation. Notably, neurons with lower Pearson correlation—those features common to both data types—tend to receive stronger inhibition (i.e., lower activation values). This supports the hypothesis that the inhibition mechanism prioritizes the suppression of non-discriminative, broadly shared signals, thereby refining the input representation presented to the router.

Figure~\ref{fig:common_vs_dis} shows the count of "common" features shared across data types versus the count of `discriminative' features for data types. For each threshold, we count how many neurons classified as "common" or "discriminative" fall below the threshold and are therefore considered suppressed. It can be seen the data scatter is mostly below the $y=x$ line, which means that the number of "common" features is more frequently suppressed. This is in line with our hypothesis that commonly shared features across data statistic types can be inhibited for the router to find the best expert using discriminatory features. 

% trivial and only state in the experiments section: random inhibition, We can apply dropout 

%\subsection{Decision in reinforcement learning} 
\section{Summary and Future Directions}
\label{sec:future}

To revisit, we have shown evidential support for our hypothesis to employ neural inhibition models for dynamic routing and Mixture-of-Expert architectures. The experimental evidence shows clear analytical and measured evidence for performance on generic tasks. Further, we present the status of background research and argue that this is an under-investigated direction. Encouraging research work for neural inhibition and dynamic shifts in deep learning architecture will not only expand methodology to advance task performance but also improve the scalability of models. 

While our current study proposes global inhibition mechanisms applied to router inputs, there are several extensions that can deepen our understanding of how inhibition can further boost the capabilities of dynamic routing. Below, we outline key directions and hypotheses that can guide future research.

\subsection{Global and Diverse Inhibition Mechanisms} 

All-to-All global inhibition mechanism is one of the most promising directions. As described in the algorithm section, All-to-All inhibition increases diversity in the entire network and extends the perspective of applying a novel technique to scalable architectures. The underlying hypothesis is that by allowing inhibition to emerge from a global view of the network’s representational space. 

\subsection{Balancing Expert Utilization} 

One persistent challenge of the current Mixture-of-Experts models is the over and under-utilization of experts. Often, multiple experts respond similarly to the same inputs, leading to inefficient use of model capacity \cite{fedus2022switchtransformersscalingtrillion, chi2022representationcollapsesparsemixture}.
A diverse neural inhibition model may play a larger role to selectively improve the representation space, and discover automated, adaptive techniques to arrive at balanced expert utilization.

\subsection{Enhancing Representation Learning} 

Neural inhibition adds methodologies to adaptively change the representation space of neuron populations. This makes the representation space richer and versatile and expands the expressiveness of representations. Further, the routing of data with MoE architectures also relies on the quality of representations. If the representation space is noisy or high-dimensional, the router’s ability to assign inputs to the correct experts is impaired. While traditional approaches have relied on increasing the capacity of encoders to resolve this \cite{Lepikhin2020, fedus2022switchtransformersscalingtrillion}, an alternative is to apply selective inhibition to the input representation before routing occurs. Further, the routing of data with MoE architectures relies on the quality of representations. The representation improves the modeling capability. While traditional approaches have relied on increasing the capacity of encoders to resolve this \cite{Lepikhin2020, fedus2022switchtransformersscalingtrillion}, there is more potential to apply selective inhibition to representations before routing occurs.

Gating mechanisms such as those in Gated Convolutional Networks \cite{dauphin2017languagemodelinggatedconvolutional} and channel gating in CNNs \cite{hua2019channelgatingneuralnetworks} demonstrate that inhibition can be applied with various deep learning architectures. For example, suppressing uninformative channels in convolutional nets improves both generalization and interpretability. Over time, this offers research opportunities to enable more efficient use of capacity and better alignment between architectures and input subdomains.

Moreover, a \textit{globally-connected inhibition} model may reduce redundancy in existing and new architectures. It selects the most discriminative signal throughout the network, which enhances representation learning of the input space. 

\subsection{Reducing Memory and Computation}
Inhibition mechanisms can also contribute to model efficiency by reducing the number of active neurons during inference. By suppressing unimportant features early in the pipeline, inhibition can reduce the computational load on subsequent layers. This is especially valuable in large-scale sparse MoE models, where expert layers are expensive. A dynamic inhibition network can temporarily turn off pathways that contribute little to the current input. It allows for sparsity while maintaining full model capacity and only activating what is needed. 

Existing work in MoE with multi-modal networks has shown promise in scaling models efficiently while leveraging modality-specific experts \cite{shen2024momemixturemultimodalexperts, lin2024uni_moe,NEURIPS2024_7d62a85e,NEURIPS2024_009729d2}. Large multi-modal networks may apply neural inhibition to improve network sparsity, reducing memory and computational costs.

\subsection{Interpretation and Visualization for Neural Inhibition} 

While we focus on inhibition on general dynamic routing tasks, we encourage future research to develop enhanced interpretation techniques that can help unveil the underlying mechanisms. Understanding the role of global inhibition in deep learning can help design algorithms to automatically improve models. Visualizing inhibition masks, activation suppression patterns, and expert selection decisions could provide insight into how inhibition affects information flow within the network. Importantly, we encourage high-value, advanced tools for visualizing inhibition in larger foundational models.

\section{Conclusion}
\label{sec:conclusion}

We propose that a biologically inspired global inhibition mechanism is critical for improving dynamic routing in Mixture-of-Experts (MoE) architectures. We verified with supporting evidence that the selective suppression of neural activations throughout the model can enhance dynamic routing decisions. We develop a research design and framework of analysis to investigate and illustrate the idea. Importantly, through empirical analysis of synthetic mixed-numbers data and language corpus data, we demonstrate that the global inhibition mechanism helps model performance on general tasks such as vision classification and word prediction tasks. We conclude that global inhibition indeed improves MoE model performance. Moreover, due to the multiple challenges current MoE models face and the merits of ``\textit{inhibition gating}'', we call for research directions and attention to bridge gaps and advance the two fields.

\bibliography{refs}
% References follow the acknowledgments in the camera-ready paper. Use unnumbered first-level heading for
% the references. Any choice of citation style is acceptable as long as you are
% consistent. It is permissible to reduce the font size to \verb+small+ (9 point)
% when listing the references.
% Note that the Reference section does not count towards the page limit.
\medskip

%%%%%%%%%%%%%%%%%%%%%%%%%%%%%%%%%%%%%%%%%%%%%%%%%%%%%%%%%%%%

\appendix

%%%%%%%%%%%%%%%%%%%%%%%%%%%%%%%%%%%%%%%%%%%%%%%%%%%%%%%%%%%%

\end{document}